%% file: commaqa.tex
\pdfoutput=1

\documentclass[11pt]{article}

\usepackage{acl}

\usepackage{times}
\usepackage{latexsym}

\usepackage[T1]{fontenc}

\usepackage[utf8]{inputenc}

\usepackage{microtype}

\usepackage{graphicx}
\usepackage{arydshln}

\usepackage{amssymb}
\include{macros}

\usepackage{listings}
\lstnewenvironment{tabularlstlisting}[1][]
 {%
  \lstset{aboveskip=-1.3ex,belowskip=-3.5ex,#1}%
 }
 {}
\usepackage{colortbl}
\definecolor{Gray}{gray}{0.9}
\definecolor{beaublue}{rgb}{0.74, 0.83, 0.9}
\usepackage{soul}

\title{Hey AI, Can You Solve Complex Tasks by Talking to Agents?}

\author{
  Tushar Khot \ \ \ \
    Kyle Richardson \ \ \ \ 
    Daniel Khashabi \ \ \ \ 
    Ashish Sabharwal\\
    \hspace{1ex} \\
 Allen Institute for AI,
 Seattle, WA, U.S.A.\\
 {\tt \small \{tushark,kyler,danielk,ashishs\}@allenai.org}\\
}

\begin{document}

\maketitle

\begin{abstract}
Training giant models from scratch for each complex task is resource- and data-inefficient. To help develop models that can leverage existing systems, we propose a new challenge: Learning to solve complex tasks by communicating with existing agents (or models) in natural language. We design a synthetic benchmark, \dataset, with three complex reasoning tasks (explicit, implicit, numeric) designed to be solved by communicating with existing QA agents. For instance, using text and table QA agents to answer questions such as "Who had the longest javelin throw from USA?". We show that black-box models struggle to learn this task from scratch (accuracy under 50\%) even with access to each agent's knowledge and gold facts supervision. In contrast, models that learn to communicate with agents outperform black-box models, reaching scores of 100\% when given gold decomposition supervision. However, we show that the challenge of learning to solve complex tasks by communicating with existing agents \emph{without relying on any auxiliary supervision or data} still remains highly elusive. We release \dataset, along with a compositional generalization test split, to advance research in this direction.\footnote{
\url{https://github.com/allenai/commaqa}
}
\end{abstract}

\section{Introduction}
\label{sec:intro}

A common research avenue pursued these days is to train monolithic language models with billions of parameters~\cite{gpt2,raffel2020exploring,gpt3} to solve every language understanding and reasoning challenge~\cite{glue,superglue}.
In contrast, humans often tackle complex tasks by breaking them down into simpler sub-tasks, and solving these by interacting with other people or automated agents whose skill-sets we are familiar with. This approach allows us to learn to solve new complex tasks quickly and effectively, by building upon what's already known.
Can AI systems learn to do the same?

\begin{figure}[t]
    \centering
    \includegraphics[width=0.4\textwidth,trim=0.5cm 0.3cm 0cm 0.3cm]{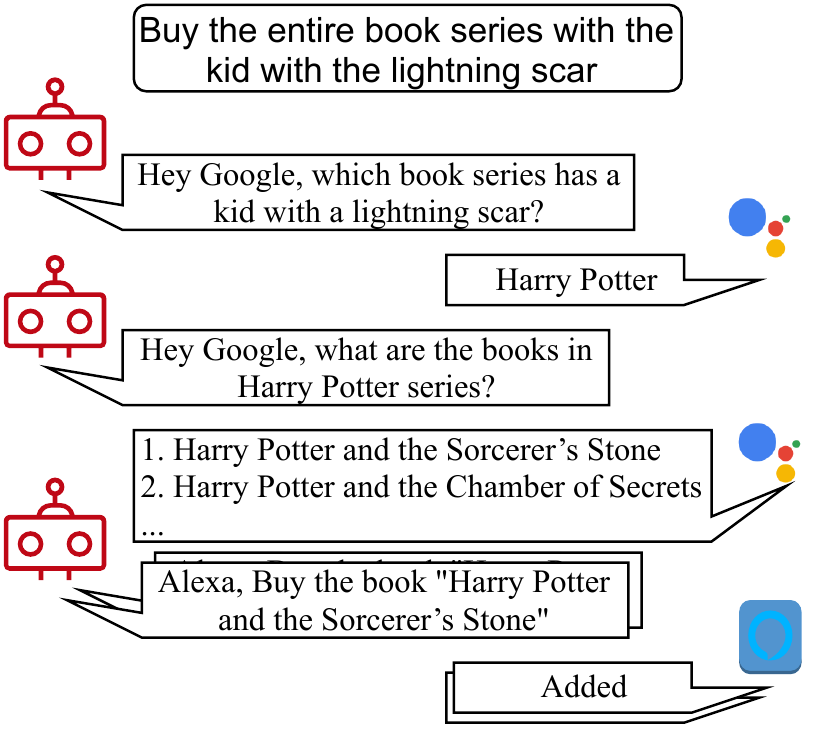}
    \caption{
      Motivating example for a setup where a system is expected to learn to accomplish goals by interacting with agents via a natural language interface.
    }
    \label{fig:intro}
\end{figure}

To facilitate research in this direction, we propose a new reasoning challenge and a benchmark called \dataset\ where, in addition to the usual end-task supervision, one has access to a set of pre-defined AI agents with examples of their natural language inputs.
Importantly, the target end-task is designed to be too difficult for current models to learn based only on end-task supervision. The goal is instead to build models that learn to solve the target task by decomposing it into sub-tasks solvable by these agents, and interacting with these agents in natural language to do so.

As a motivating example, consider the interaction depicted in Figure~\ref{fig:intro} where a system is asked to buy a book series with a certain property. The system breaks this goal down, using agent-1 (here Google Assistant) to identify the referenced book series as well as the list of books in that series, and then using agent-2 (here Amazon Alexa) to make the purchase. While both of these agents interact with the system in natural language, they have different and complementary skill sets,\footnote{but not necessarily mutually exclusive skills} rely on privately held knowledge sources, and have been built at an enormous cost. At the same time, neither agent by itself can accomplish the original goal.

An alternative to building such a system that interacts with existing agents is to teach all requisite sub-tasks and skills to a large black-box system, say via multi-task learning~\cite{2020unifiedqa,Gupta2021TowardsGP}. This, however, not only wastes time and resources, but is often also infeasible. For example, agents such as Google Assistant and OpenAI GPT-3 use private knowledge resources and are computationally expensive to train even once. It would thus be nearly impossible to build a single system with the capabilities of both of these agents. 

We note that agents need not be sophisticated AI assistants. An agent may simply be a previously developed question-answering (QA) model, a math module, a function of textual input, an image captioning system---anything the community already knows how to build. The goal is to \emph{learn to leverage existing agents for more complex tasks.}

To enable the development of general systems for this task, we identify the minimal inputs that must be assumed for the task to be learnable---training data for the complex task, existing agents that together can solve the complex task, and examples of valid questions that can be asked of these agents (capturing the agents' capabilities). We build a new synthetic benchmark dataset called \dataset\ (\underline{Comm}unicating with \underline{a}gents for \underline{QA}), containing three complex multihop QA tasks (involving \underline{E}xplicit, \underline{I}mplicit, and \underline{N}umeric reasoning) and four input QA agents that can solve these tasks.

\dataset\ is not yet another multi-hop reading comprehension dataset. It is designed to facilitate the development of a new family of techniques that teach systems to communicate with a wide variety of agents to solve different types of complex tasks.

We demonstrate that black-box models struggle on \dataset\ even when provided with auxiliary data, such as domain-relevant agent knowledge. On the other hand, a model that leverages the agents~\cite{khot-etal-2021-text} can achieve very high accuracy but relies on auxiliary supervision (decomposition annotations). While it is possible to identify valid decompositions using just the end-task labels, the search space is extremely large and na\"{i}ve approaches, as we show, help only with one of the datasets. \dataset\ thus serves as a new challenge for the NLP community.

\textbf{Contributions:} We (1) propose a new challenge of learning to solve complex tasks by communicating with agents; (2) develop a synthetic multi-hop QA dataset \dataset\ with three reasoning types; (3) provide auxiliary training data and a compositional generalization test set; (4) demonstrate the challenging nature of \dataset\ for black-box models; and (5) show the promise of compositional models that learn to communicate with agents.

\section{Related Work}
\label{sec:related-work}

\textbf{Multi-hop QA}~\cite{MultiRC2018,obqa,qasc,geva2021strategyqa} focuses on reasoning with multiple facts. Some multi-hop datasets~\cite{hotpotqa,Dua2019DROP} have been used to develop modular approaches such as TMNs~\cite{khot-etal-2021-text}, which are a step towards our goal---they try to solve complex questions by leveraging agents such as single-hop QA models. However, these approaches have had limited success because current datasets are insufficient for the development of such models, for two reasons. First, prevalent single-hop shortcuts~\cite{min2019compositional,Trivedi2020DiRe} incentivize models trained on answer supervision alone to learn to exploit these shortcuts rather than learn to compositionally communicate with agents. E.g., they learn to answer a multi-hop question by just asking one single-hop question~\cite{decomprc}. Second, these datasets often contain sub-problems not solvable by existing models/agents, such as producing structured output (e.g., outputting a \emph{list} of all touchdowns mentioned in the context).\footnote{For instance, 65\% of the errors of the ModularQA system~\cite{khot-etal-2021-text} on HotpotQA were due to questions unanswerable by existing agents. Hence these datasets don't satisfy the basic task requirement of being solvable using existing agents. This makes the learning-to-communicate task ill-defined over these datasets and meaningful progress infeasible.}

\textbf{Semantic Parsing} typically focuses on mapping language problems to executable symbolic representation based on a pre-defined grammar~\cite{Krishnamurthy2017NeuralSP,Chen2020Neural}. Similar ideas are also found in the area of program synthesis~\cite{gulwani2011automating,desai2016program}. 
These goals, like ours, seek to simplify complex problems into simpler executable forms, without relying on explicit intermediate annotation~\cite{clarke2010driving,berant2013semantic}. We, however, diverge from this line by seeking agent communication in free-form language, not bound to any pre-specified set of operations or domain specific languages.

\textbf{Question Decomposition} is used to solve multi-hop QA but the resulting models~\cite{talmor2018web,decomprc,perez2020unsupervised,khot-etal-2021-text} are often dataset-specific, rely on decomposition annotations, and limited to one or two QA agents. To address these limitations, our proposed challenge covers three dataset types and four agents. Additionally, models are expected to learn to decompose the task by interacting with the agents, rather than relying on human annotations.

\textbf{Synthetic Reasoning Challenges} 
have recently been proposed~\cite{lake2018generalization,sinha2019clutrr,clark2020transformers,betz2021deepa2} to help systematically identify the weaknesses of existing models and inspire modeling innovation~\cite{liu2021can}. Our new tasks are unique and focus on simulating complex agent interaction to motivate the development of decomposition-based modeling approaches.

\textbf{Text-Based Games}, similar to our work, involve interacting in plain text in order to accomplish a goal~\cite{yuan2019interactive,yuan2020interactive,hausknecht2020interactive,ammanabrolu2021motivate,jansen2021systematic}.  This is typically done in a physical environment, which acts as an ``agent'' in our setting. Unlike many works in this area, we focus on different classes of compositional questions (e.g, implicit, numerical) and formulate a challenge that makes minimal assumptions about having access to agents' internal information or input language.

\section{Challenge Task Definition}
\label{sec:learning-to-talk-with-agents}

We formalize the new challenge task of \emph{learning to talk with agents to solve complex tasks}. To ensure generality of solutions, we identify minimal inputs for the task to be well-defined and learnable. 

First we must define $\{\subm_i\}_{i=1}^m$, the agents or models that solve simpler sub-tasks.\footnote{As mentioned earlier, we use \emph{agents} to refer interchangeably to models, assistants, or functions that take free-text as input and produce free-text as output.} Minimally, we need to define the space of valid inputs $\subml_i$ for each agent $\subm_i$, i.e., how can they be invoked. For a system to identify the appropriate agent for each sub-task, we also need to define the capabilities of each agent. Since these agents are often defined for natural language tasks, the space of inputs captures the capabilities of these agents too.  For instance, "Buy the book `Harry Potter and the Sorcerer's Stone'"
captures the Alexa agent's capability of buying books. Instead of complex formal specifications of the agent's capabilities, we use natural language inputs as a rich and convenient representation.

Next, we need a target task $\complext$ that can be solved via a composition of the capabilities of various $\subm_i$.\footnote{Existing datasets lack this requirement, making it impossible to focus only on the agent communication aspect.} Finally, to pose this as a machine learning problem, we need training data $\complexd = \{(\cx_k, \cy_k)\}_{k=1}^{N}$ for $\complext$. Since collecting annotations for complex tasks can be difficult, $\complexd$ is expected to be relatively small. Models must therefore use the available agents, instead of learning the complex task from scratch.

Given these pre-requisites, we can define the challenge task
as follows:

\textbox{
\textbf{Challenge}: Learn a model to solve a complex task $\complext$, given only:\\
- Training dataset $\complexd = \{(\cx_k, \cy_k)\}_{k=1}^{N}$ for $\complext$; \\
- Agents $\{\subm_1, \ldots, \subm_m\}$ that can help solve $\complext$; \\
- Examples from the space $\subml_i$ of valid inputs for each agent $\subm_i$ that captures its capabilities.
}\\

One example of this challenge is answering multi-hop questions given two agents: an open-domain TextQA agent $\subm_1$ and an open-domain TableQA agent $\subm_2$. Agent $\subm_1$ can use large textual corpora to answer questions such as "Who directed Kill Bill?". Agent $\subm_2$ can use tables (e.g., Filmography tables) to answer questions such as "List the movies directed by Quentin Tarantino".  Finally, the training data $\complext$ for the complex task would contain examples such as
("What movies has the director of Kill Bill appeared in?", ["Reservoir Dogs", ...,]).

\begin{figure*}[tb]
    \centering
    \includegraphics[scale=0.6,trim=0cm 0.3cm 0cm 0.6cm]{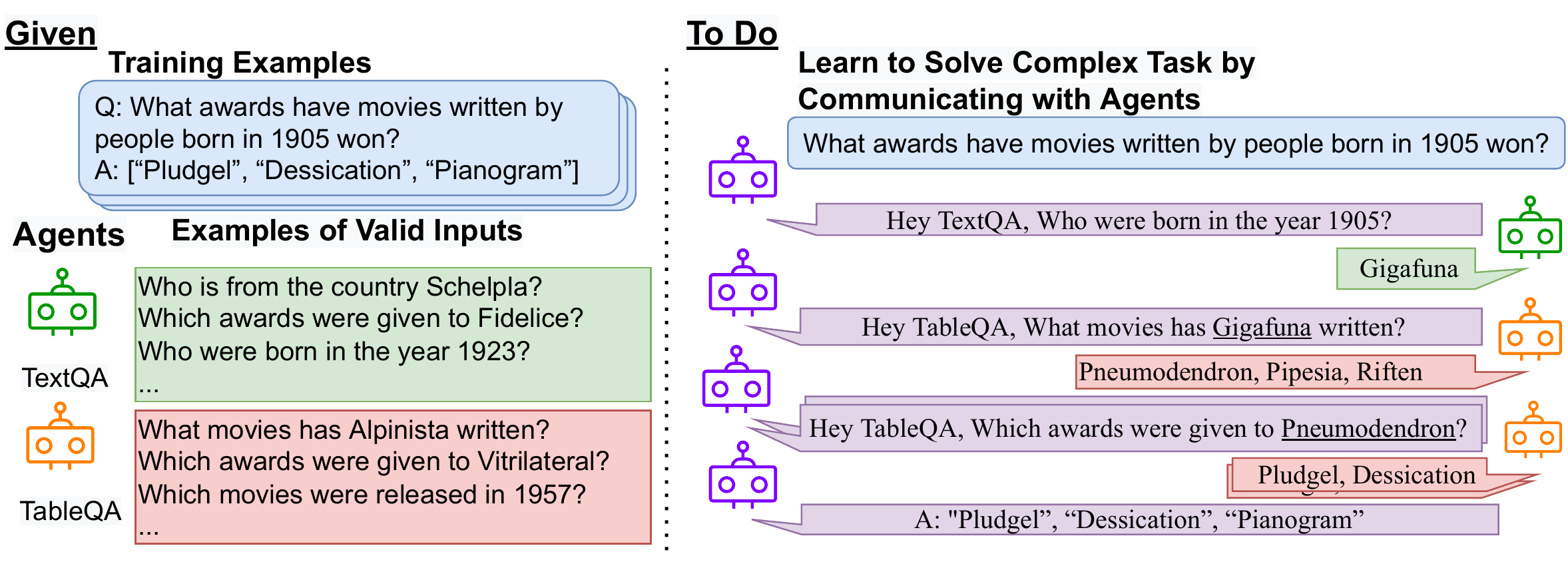}
    \caption{High-level overview of the task, with examples from \datasete. Given the agents, their valid inputs, and training examples for a complex task, the goal is to learn to solve this task by communicating with the agents.}
    \label{fig:task_defn}
    \vspace{-2ex}

\end{figure*}

\section{Dataset: \dataset\ Benchmark}
\label{sec:dataset}

We next propose a new benchmark dataset \dataset\ that enables the development of models that can learn to communicate with existing agents. Specifically, we provide a collection of \emph{three synthetic datasets} where each question is answerable by talking to simple QA agents. Note that we are not proposing a new class of questions but a new dataset for the proposed challenge task. A high-level overview of this dataset is shown in Fig.~\ref{fig:task_defn}.

We choose QA as the underlying task and use QA agents for this challenge because the question-answer format can capture a broad range of tasks~\cite{Gardner2019QuestionAI} while also naturally surfacing the capability of each agent. 
For instance, the question "What are the key frames in v?" describes a capability of the invoked agent (namely, identifying key frames), in addition to the specific inputs. We next describe our framework for building \dataset, which we believe can be extended to other complex tasks, e.g., video summarization.

\subsection{Agent Definition}

To define the $i$-th agent, we build a knowledge base that captures its internal knowledge resource $\kb_i$. We use natural language question templates to define the set of questions that this agent can answer over this internal knowledge. For example, given a KB with relations such as "directed(x, y)", the agent would answer questions based on the template: "Who directed the movie \_\_?"

\paragraph{Knowledge Base, $\kb_i$.}
To build the knowledge base, we define a KB schema as a set of binary relations between entity types, e.g., director(movie, person). We build a list of entity names that belong to each entity type. To avoid potential conflicts with the LM's pre-training knowledge, all entity names are generated non-existent words.\footnote{\url{https://www.thisworddoesnotexist.com}}

Rather than building a static and very large KB, we sample \emph{a possible world} independently for each question, by sub-sampling entities for each entity type and then randomly assigning the KB relations between these entities. This prevents memorization of facts across the train and test splits, which in the past has led to over-estimation of QA model performance~\cite{Lewis2021TrainTest}. This also encourages models to learn proper multi-hop reasoning using the agents, rather than memorizing answers. 

\paragraph{Examples of Valid Inputs.}
To define the space of valid inputs for each agent $\subm_i$, we define a set of question templates that can be answered by it over $\kb_{ik}$ (e.g., Who directed \_\_?).
We construct questions corresponding to a relation in both directions, e.g., "Who all directed \_\_?" and "For which movies was \_\_ a director?". To emulate diversity in natural language, we specify multiple phrasings for the same question. We use these templates to generate examples of valid inputs in $\subml_i$ by grounding them with entities of the appropriate entity type (e.g., Who directed Kill Bill?).
 
To ensure generalization to a broad set of tasks, we do not limit the questions to only single span answers. Depending on the question, the agent can produce answers as a single string (span, boolean or a number), a list of strings (e.g., "Which movies did Spielberg direct?"), or a map (e.g., "What are the states and their capitals in USA?"). 

\paragraph{Implementation.}
To answer the question, agents convert questions into queries against their internal knowledge (based on the templates) which we implement as  a symbolic function (written in Python), instead of a model. While a language model might be able to generalize to out-of-distribution variations in language, its behavior can be often unpredictable. By implementing the agents as pattern-based functions, we ensure that the resulting systems would stay within the language constraints of each agent and generalize to restricted language models. Additionally, this enables faster development of approaches without spending resources on running a large-scale LM for each agent. 

\begin{table*}[tb]
    \centering
    \includegraphics[width=0.95\textwidth,trim=0cm 0.3cm 0cm 0.5cm]{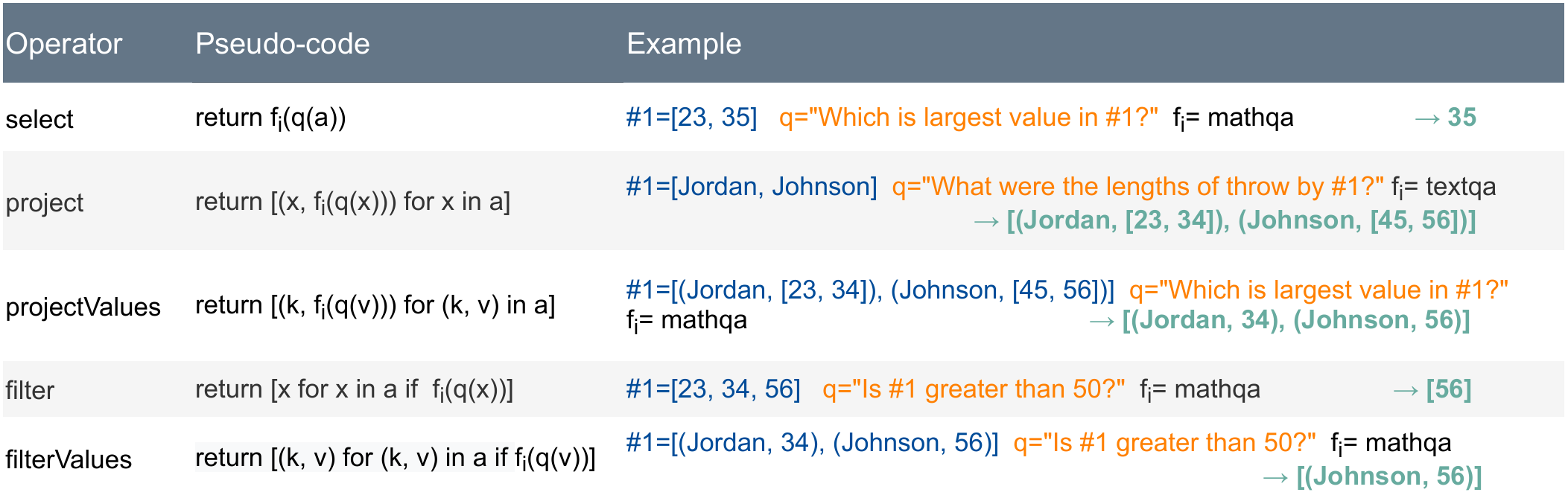}
    \caption{Compositional Operators used in this work to transform structured answers into queries answerable by an agent. The operator takes the agent $\subm_i$, a structured answer $a$ (we use the answer index, e.g., \#1, to refer to any answer), and a query with a placeholder as inputs and executes the pseudo-code shown here. 
    }
    \label{fig:operations}
\end{table*}
\subsection{Complex Task Definition}
Given the space of valid input questions for each agent, we construct training examples for the complex task using templated theories. These theories consist of a complex question template and a composition rule expressed as a sequence of questions asked to appropriate agents. For example, 
\textbox{
\small
"What movies have the directors from \$1 directed?"\\
 $\#1$ = [textqa] "Who is from the country \$1?"\\
 $\#2$ = [tableqa] "Which movies has \#1 directed?"
}

\paragraph{Composition Operators.} 
While this simple theory would work for single span answers, these agents often return list or map answers. Even within this simple example, there can be multiple directors from a given country and this list cannot be directly fed to the tableqa model, i.e., "Which movies has [...] directed?". This problem gets even more challenging with complex structures. E.g., maintaining a map structure while operating on the values of the map (see 3rd row in Table~\ref{fig:operations}).

To handle the different answer structures, we define a special set of compositional operators in Table~\ref{fig:operations}. These operators take agent $\subm_i$, a structured answer $a$, and a query with a placeholder as inputs, and execute a set of queries (as defined by the pseudo-code in Table~\ref{fig:operations}) against $\subm_i$.
These operators are inspired by QDMR~\cite{Wolfson2020Break}, but modified to be actually \emph{executable}. E.g., the "project" operator in QDMR: "return directors of \#1?" does not specify how to execute this query whereas our operation \opmodq{project}{textqa}{Who are the directors of \#1?} specifies how to use the TextQA model and \#1 to generate a map.

We also define a set of agent-independent data structure transformations in Table~\ref{tab:transformations}, e.g., convert a map into a list of its keys.  Since longer chains of reasoning are prone to more errors~\cite{Fried2015HigherorderLS,Khashabi2019OnTP}, we don't model these simple transformations as additional reasoning steps. Instead, 
we concatenate compositional operators with transformations to create about 20 new, combined operators such that transformations can be applied after an operation in a single step, e.g., project\_Values operation performs the project operation followed by the Values transformation.

Given these operators, the final theory for the above example would look like:
\textbox{
\small
"What movies have the directors from \$1 directed?"\\
 $\#1$ = \opmodq{select}{textqa}{Who is from the country \$1?}\\
 $\#2$ = \opmodq{project\_values\_flat\_unique}{tableqa}{Which movies has \#1 directed?}
}

\begin{table}[htb]
    
    \small
    \centering
    \begin{tabular}{lp{5.5cm}}
    Transf. & Procedure \\
    \hline
    \textsc{Flat}     & Flatten list of lists into a single list \\
    \textsc{Unique}     & Return the unique items from a list \\
    \textsc{Keys} &  Return the list of keys from a map \\
    \textsc{Values} & Return the list of values from a map 
    \end{tabular}
    \caption{Simple transformations that modify the output data structure. These transformations can be chained together with an operation, e.g., \textsc{Project\_Values}.}
    \label{tab:transformations}
    \vspace{-2ex}
\end{table}

\begin{figure*}
    \centering
    \includegraphics[scale=0.69,trim=0cm 0.5cm 0cm 1cm]{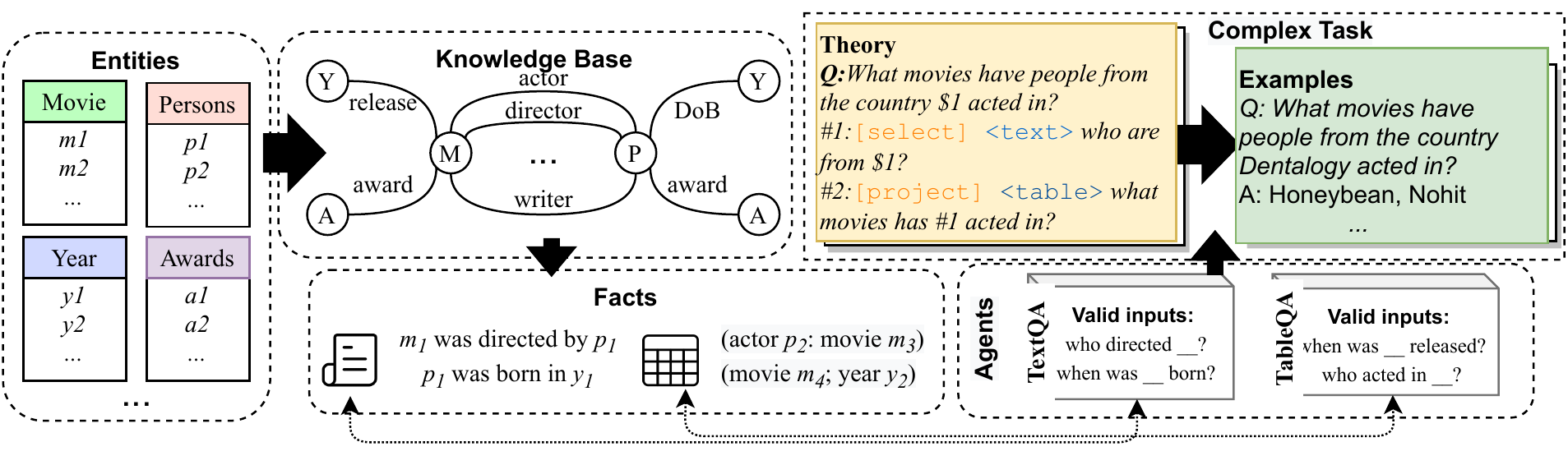}
    \caption{High-level schema of our dataset construction process. We use a list of entities and a KB schema to generate a list of facts. The QA agents operate over these facts to answer a set of pre-determined questions that form the examples of valid inputs from $\subml_i$. We define multiple complex question templates and a corresponding \emph{theory} that can be used to answer them. We then ground these question templates (i.e. sample \$1) to create complex questions and use the agents to generate the answers.
    }
    \label{fig:construction}
    \vspace{-1ex}
\end{figure*}

\paragraph{Building Examples.}
Given a KB schema, question templates for each agent, and theories, we can now build examples for the complex task (Fig.~\ref{fig:construction}). We first sample a possible world based on the KB schema. We assign each relation to one of the agents based on which agents are likely to answer such questions, i.e., only this agent would answer questions about this relation.
 This captures multi-modality of knowledge, e.g., movie awards might be described in text or a table, but a person's birth date is likely described in text. When a relation can be captured by knowledge in multiple modalities, it is assigned to one of them per KB. This emulates the challenging setting where a model must interact with multiple agents to find the answer.\footnote{With real questions and agents, models may be able to avoid this by just memorizing the agents.} We use the templated theories to construct questions by grounding placeholders. We select $m$ valid questions\footnote{has a non-empty answer and up to five answer spans} for each KB such that each theory has the same number of examples across the dataset.

\subsection{Auxiliary Information}
In addition to the basic task definition, we also consider auxiliary information that may be available in some cases. The main goal of this information is to (a) provide stepping stones for development of methods towards the final goal of learning to communicate with agents using answer supervision only, and (b) evaluate the abilities of current state-of-the-art assuming access to this additional information. We emphasize that such auxiliary information may not always be available (e.g., when using a proprietary agents such as Alexa). 

We consider two kinds of such information---\emph{auxiliary supervision} for the complex task's training examples $(\cx_k, \cy_k) \in \complexd$, and \emph{auxiliary data} about the agents $\{f_i\}$ themselves (not tied to $\complexd$). This is summarized in Table~\ref{tab:aux-info}.

\begin{table}[ht]
\textbox{
\emph{Auxiliary Supervision for $(\cx_k, \cy_k) \in \complexd$}: \\ 
- Gold Decomposition $\decomp_k$ for $\cx_k$ \\ 
- Gold Knowledge $\fact_k$ for $\cx_k$ 
\vspace{5pt} \hrule \vspace{5pt} 
\emph{Auxiliary Data for agents $\{\subm_i\}$}: \\ 
- Training data $\submd_i = \{(\submx_{ij}, \submy_{ij})\}_{j=1}^M$ for agent $\subm_i$, where $\submx_{ij} \in \subml_i$ and $\submy_{ij} = \subm_i(\submx_{ij})$
 \\
- Complete knowledge resource $\kb_i$ used by $\subm_i$, or a manageable subset $\kb_{ik} \subset \kb_i$ containing $\fact_k$
}
\caption{Auxiliary information as stepping stones towards the full \dataset\ task.}
\label{tab:aux-info}
\end{table}

For auxiliary \underline{supervision}, we consider having access to annotated decomposition $\decomp_k$ of a complex task training input $\cx_k$ into valid inputs for various agents. We also consider annotated gold facts $\fact_k$ that could be used to answer $\cx_k$.

For auxiliary \underline{data}, we consider having access to the training data used to build the agents, or the underlying knowledge base $\kb_i$ used by them (and possibly even a question-specific relevant subset $\kb_{ik}$). For example, $\kb_i$ would be equivalent to the entire text and table corpora used by TextQA and TableQA agents, and $\kb_{ik}$ could be the texts and tables relevant to the question domain (e.g., movies). Such information can be used to train a stronger black-box model on the end-task, e.g., fine-tuning on the agent's training data first or using the gold facts to identify relevant context. These approaches that circumvent the agents are not the target of our dataset, but we nevertheless evaluate them to highlight their limits.

\begin{figure*}
    \centering
    \includegraphics[scale=0.7,trim=0.0cm 0.3cm 0cm 0.6cm]{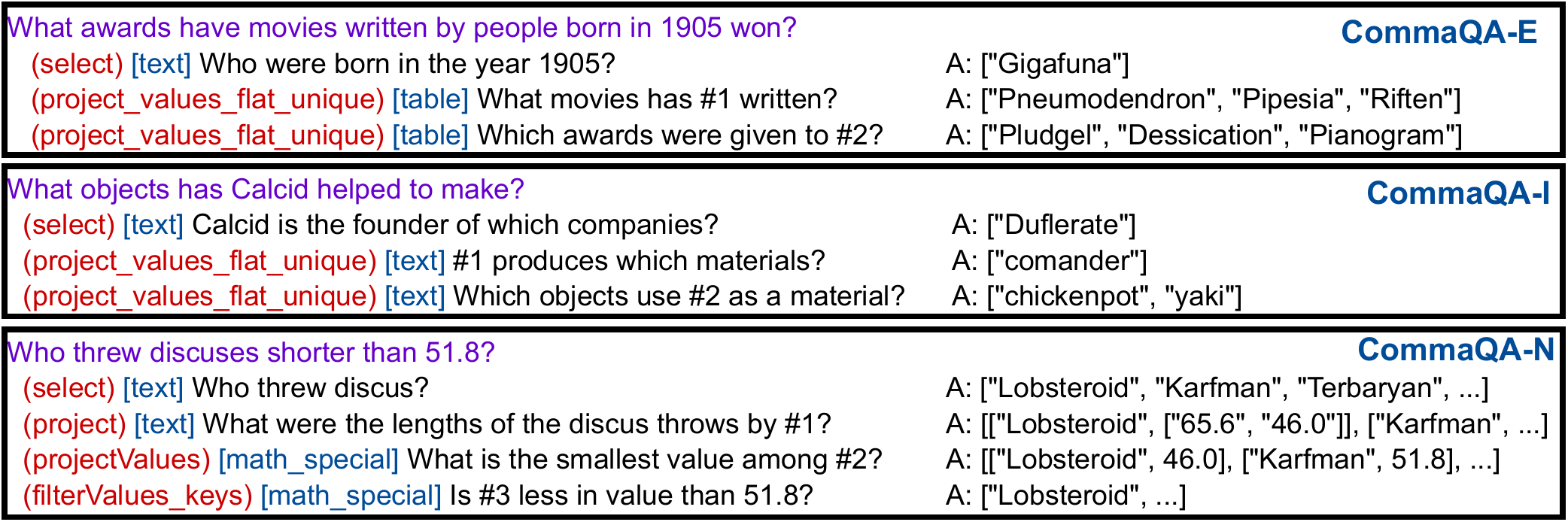}
   \caption{Sample Decomposition Annotations for example questions in \dataset. We denote the composition operators using the format \opmodq{operation}{agent}{question}. }
    \label{fig:decomp_eg}
    \vspace{-2ex}
\end{figure*}

\paragraph{Building Auxiliary Information.}
We generate the gold decomposition $\decomp_k$ for each example $\cx_k$ using the same language as the theories (see Fig.~\ref{fig:decomp_eg}). We verbalize each relation to create the underlying knowledge resource $\kb_{ik}$ used by the agent $\subm_i$ (e.g., relation director(M,  P) is converted into "M was a movie directed by P" or "movie: M ; director: P" depending on the agent assigned to this relation). While our KB and resulting facts are intentionally simple to show the limitations of black-box models, such verbalization may not always be possible with larger KBs and hence should not be relied upon. For each training example, we collect the facts used by each agent in the decomposition and treat these as gold facts $\fact_k$.

\subsection{\dataset\ Dataset}
We use the above framework to build three datasets capturing three challenges in multi-hop reasoning.

\textbf{\datasete: Explicit Decomposition.}
This dataset consists of multi-hop questions from the movie domain where the reasoning needed to answer the question is \textbf{E}xplicitly described in the question itself~\cite{hotpotqa,xanh2020_2wikimultihop,musique}. For example, "What awards have the movies directed by Spielberg won?". We use a TextQA and TableQA agent where certain relations can either be expressed in text or table (more details in App. Fig.~\ref{fig:datasete}).  

\textbf{\dataseti: Implicit Decomposition.}
This dataset consists of multi-hop questions where the reasoning needed is \textbf{I}mplicit~\cite{qasc,geva2021strategyqa}, for example, "Did Aristotle use a laptop?". Inspired by such questions in StrategyQA~\cite{geva2021strategyqa}, we create this dataset using three agents(TextQA, KBQA and MathQA) with just two question styles: (1) "What objects has \_\_ likely used?" and (2) "What objects has \_\_ helped make?". However each question has three possible strategies depending on the context (see App. Fig.~\ref{fig:dataseti} for more details). This is a deliberate choice as similar sounding questions can have very different strategies in a real world setting, e.g., "Did Steve Jobs help develop an Iphone?"  vs. "Did Edison help develop the television?".

\textbf{\datasetn: Numeric Decomposition.}
This dataset consists of \textbf{N}umeric (also referred to as discrete) reasoning questions~\cite{Dua2019DROP,Amini2019MathQATI} requiring some mathematical operation, in addition to standard reasoning. For example, "Who threw javelins longer than 5 yards?". We create this dataset in the sports domain with TextQA, TableQA and MathQA agents (more details in App. Fig.~\ref{fig:datasetn}).

\paragraph{Dataset Statistics.}
The final dataset\footnote{released under CC BY license} consists of the three QA sub-datasets described above, key statistics summarized in Table~\ref{tab:dataset_stats}.

\begin{table}[ht]
    \centering
    \small
    \begin{tabular}{l|rrr}
         & \multicolumn{3}{c}{\dataset} \\
         & E & I & N \\
        \hline
         \T \#questions &          10K & 10K & 10K \\
         \#theories & 6 & 6 & 6 \\
         \#steps per theory & 2.7 & 3.2 & 4.7 \\
         \#entity types & 7 & 13 & 5 \\
         \#relations & 11 & 16 & 4 \\
         \#templates in $\subml_i$ & 42  & 68  & 30 \\
         \#entities per answer & 3.21 & 3.29 & 1.36 \\
         \#KB facts per KB & 169.4 & 175.7 & 80 \\
         \#T5tokens per KB & 2252.9 & 2540.9 & 1513.4 \\
         \#Gold facts per qn & 7.5 & 6.9 & 15.4 \\
    \end{tabular}
    \caption{Statistics of \dataset. All per-question and per-KB statistics are averages.}
    \label{tab:dataset_stats}
    \vspace{-2ex}
\end{table}

There are 10K total examples in each dataset with 80\%/10\%/10\% train/dev/test split. To prevent models from guessing answer spans, we introduce more distractors by sampling a large number of facts for \datasete\ and \dataseti. 

This results in a larger number of facts in the KB ($\sim$170) and larger length of the KB in these two datasets($\sim$2500 tokens). Since \datasetn\ can have derived answers from numeric reasoning and has longer chains (avg \#steps 4.7 vs.\ 2.7 in \datasete), we do not need a large number of distractor facts (80 facts/KB).

\paragraph{Metrics.}
The answer $y_k$ to each question $x_k$ in \dataset\ is an unordered list of single-word entities.\footnote{Although not in the current dataset, entities in the unordered list $y_k$ may be repeated, i.e., we have a multi-set.} By the design of the dataset, a model that performs the desired reasoning should be able to output $y_k$ correctly, barring entity permutation. Hence, we use \emph{exact match accuracy} as the metric.\footnote{Our implementation uses "exact match" in the DROP multi-span evaluator, which accounts for entity reordering.} (see appendix for a softer metric, F1 score)

\section{Experiments}

We evaluate various models on \dataset, including a baseline model (with no auxiliary information) for the task and state-of-the-art models that have access to auxiliary information.

\subsection{Models}

\subsubsection{\dataset\ Baseline Model}
We develop a baseline approach that directly targets the challenge task without relying on any auxiliary information. Specifically, we use the Text Modular Network (TMN) framework~\cite{khot-etal-2021-text} that trains a \texttt{NextGen} model that communicates with the agents. This model is trained to produce the next question (including operation and agent) in a decomposition chain, given the questions and answers so far, which is then executed against the agent to produce the answer for the current step. Additionally this framework samples multiple questions at each step of the chain to search\footnote{Score is the sum log likelihood of the generated questions.} for the most likely chain of reasoning.

We generate the training data for \texttt{NextGen} via distant supervision. Specifically, we perform a 
na\"ive brute-force search where we sample $l$ questions at each step for up to $o$ steps.\footnote{$o$ is set based on the length of the rules in each dataset, i.e., $o=3$ for \datasete, $o=4$ for I, $o=7$ for N.} The operations are chosen randomly but we only consider the applicable operations (e.g., "select" for the first step). We use lexical overlap between the questions in the examples of valid inputs and the complex question to avoid wasteful random sampling.\footnote{We also found random generally performed worse.} We assume all chains that lead to the gold answer\footnote{We use exact match since the correct decomposition with our error-free agents should lead to exactly the gold answer.} represent valid decompositions, and use them to build the training dataset for TMNs. We refer to the model as TMN-S$_l$ (see App.~\ref{app:search_model} for details).

\subsubsection{Auxiliary Supervision Models}
We next present models that depend on auxiliary information and hence target a simpler variant of the task: (1) a model trained to communicate with agents using gold decomposition supervision, \decomp;  (2) a black-box model trained to answer questions given all the agents' knowledge, $\kb_{i}$; and (3) a two-stage model that first identifies the most-relevant context (using gold knowledge supervision $\fact_{i}$) and uses this shorter context to answer the question.

\textbf{Models with Decomposition Supervision:}
Given decomposition supervision, we can directly use this gold data to train the  \texttt{NextGen} model. We refer to this model as TMN-S when we use this \textbf{s}earch and TMN-G when we \textbf{g}reedily select the most likely question at each step.

\textbf{Models with Access to Agent Knowledge:}
Given access to the facts associated with each (train or test) question $\cx_k$, i.e., \emph{each agent's domain-relevant knowledge $\kb_{ik}$}, the facts can be concatenated to create a context and frame the challenge as a reading comprehension (RC) task.\footnote{\label{footnote:not-the-purpose-of-the-dataset}We reiterate that it is often unreasonable to expect access to $\kb_i$ and especially $\kb_{ik}$. This model tries to solve \dataset\ without invoking agents, which deviates from the purpose of our benchmark dataset. Nevertheless, we conduct experiments in this setting for completeness.} We train two standard black-box models, T5-L \cite{raffel2020exploring} and UnifiedQA-L~\cite{2020unifiedqa},\footnote{We use T5 models as they can handle longer contexts.} to generate answers\footnote{We alphabetically sort answers for a deterministic order.} given a question and context. 

\textbf{Models with Fact Supervision:}
If, in addition to access to the underlying knowledge $\kb_{ik}$, we also have  the auxiliary supervision for the gold facts $\fact_k$, we can use this annotation to train a model to first retrieve a small subset of relevant facts from $\kb_{ik}$ (see App.~\ref{app:model_fact} for details). Since the context is shorter, we also train a T5-3B model\footnote{T5-11B performed worse than or same as the 3B model.} on this task.

\begin{table}[tb]
    \centering
    \small
    \begin{tabular}{ll|ccc|c}
    Model  & Aux. Info &  E & I & N & Avg. \\
    \hline
    \selrow \T TMN-S$_{5}$  &  & \phantom{0}0.0  & 0.0* &  \phantom{0}0.0 &  \phantom{0}0.0 \\
    \selrow \T TMN-S$_{10}$  &  & 17.0  & 0.0* &  \phantom{0}0.0 & \phantom{0}5.7\\
    \multicolumn{6}{l}{\underline{Auxiliary Supervision Models}}  \\ 
    \T T5-L    & $\{\kb_{ik}\}$ &  \phantom{0}0.9 & 10.2 &  35.4 & 15.5 \\     
    \T UQA-L    & $\{\kb_{ik}\}$ &  \phantom{0}1.0 & 10.2 & 39.0 & 16.7 \\ \cdashline{1-6}
    \T T5-L  & $\fact_k$, $\{\kb_{ik}\}$ & 42.2  & 49.4 & 44.7 &  45.4 \\
    \T UQA-L & $\fact_k$,$\{\kb_{ik}\}$ & 40.1 & 49.7 & 43.4  & 44.4 \\
    \T T5-3B      & $\fact_k$, $\{\kb_{ik}\}$ & 42.3 & 49.9 & 43.4 & 46.2 \\ \cdashline{1-6}
    \T TMN-G & $\decomp_k$ & 75.4 & 36.0 & \kern-1ex 100.0  & 70.5 \\     
    \T TMN-S & $\decomp_k$ & \kern-1ex 100.0 & \kern-1ex 100.0 & \kern-1ex 100.0 &  \kern-1ex 100.0 
    \\ \hline
    \end{tabular}
    \caption{Accuracy of models trained and tested separately on the 3 datasets. Last column reports average accuracy across the datasets (weighed equally). 
    \textbf{TOP} highlighted rows: Target models for \dataset\ that try solve the task using no auxiliary supervision by communicating with agents. Naive search is able to generate some training data for \datasete\ but does not result in any valid decomposition (indicated by $*$) on \dataseti. 
    \textbf{BOTTOM} rows: Models that rely on auxiliary supervision. Black-box models struggle even when given the domain-relevant KB $\kb_{ik}$. Using the additional fact supervision $\fact_k$ helps these models, but their accuracy remains below 50\%. TMN models with auxiliary decomposition supervision $\decomp_k$ can solve all tasks with search ("TMN-S") indicating that the task is solvable by communicating with agents.
    }
    \label{tab:results}
\end{table}

\subsection{Results}

Table~\ref{tab:results} reports the accuracy of these four classes of models on the \dataset\ dataset.

\textbf{Baseline model has near-zero accuracy:}
The top two rows represent baseline models that use brute-force search to generate training data for TMNs. For \dataseti, we don't find even a single chain leading to the gold answer, and hence no training data. With \datasete\ and \datasetn, we do find valid decompositions for a subset of the questions (see statistics in Table~\ref{tab:search_stats} of Appendix), but not enough to train an effective \texttt{NextGen} model. Expanding the search to $l$=20 helps achieve $\sim$100\% accuracy on \datasete\ (with $\sim$700K agent calls). However, we don't observe any gains on \dataseti\ and \datasetn\ with even 2M agent calls (see App.~\ref{app:costvacc}).

\textbf{Black-box models struggle even with access to agent knowledge:} Due to the large number of distractors, black-box models --- even with access to agent knowledge at both train and test time --- struggle to learn the task across all three datasets with average accuracy below 20. The extremely low performance on \datasete\ is especially notable, given that the reasoning needed for each question is explicitly described. While these models are able to solve similar datasets~\cite{hotpotqa}, the low scores on our synthetic dataset with more distractors indicates that they are still unable to truly learn this kind of reasoning.

\textbf{Fact annotations help but are insufficient:} Models trained on shorter context (obtained by relying on gold fact training annotation) are able to take advantage of the reduced number of distractors, improving their score to about 45 pts across all datasets. However, even with the larger 3B model, there is no noticeable improvement, indicating 45 pts being roughly a ceiling for these models.

\textbf{\dataset\ is solvable by talking to the agents:}
The TMN model, if given gold decomposition annotation for training, can solve this task (bottom two rows). This experiment is an oracle setting that shows that \dataset\ is noise-free, unambiguous, and solvable by a model that learns to talk to the agents (as designed). Note that greedily selecting the next question results in much lower performance on the two datasets (E and I) that have multiple decompositions for the same question.

\subsection{Compositional Generalization}

\newcommand{\cg}{$^\mathrm{CG}$}

We also design compositional generalization test sets \datasete\cg\ and \datasetn\cg. Specifically we create questions using novel composition of queries that have been seen during training but never together in this form. For instance, we create a new question "What awards have the directors of the \_\_ winning movies received?", given that the model was trained on questions such as "What awards have the actors of the \_\_ winning movies received?", "What movies have the directors from \_\_ directed?", and "What movies have people from the country \_\_ acted in?". 

\begin{table}[t]
    \centering
    \small
    \setlength\tabcolsep{10pt}
    \begin{tabular}{lc|cc}
    Model & Aux.~Info & E\cg  & N\cg  \\
    \hline
    \selrow \T TMN-S$_{10}$ &  & 16.2 & \phantom{0}0.0 \\
    \multicolumn{4}{l}{\underline{Auxiliary Supervision Models}}  \\ 
    \T T5-L & $\fact_k, \{\kb_{ik}\}$ & 37.0  & \phantom{0}2.0 \\ 
    \T T5-3B & $\fact_k, \{\kb_{ik}\}$ & 39.2  & 23.8  \\   \cdashline{1-4}
    \T TMN-S & $\decomp_k$ & 79.4  & 97.6 \\ \hline
    \end{tabular}
    \caption{Lower accuracy on compositional generalization test sets. TMN-S with decomposition supervision still outperforms other models.}
    \label{tab:comp_gen}
    \vspace{-2ex}
\end{table}

As shown in Table~\ref{tab:comp_gen}, all models exhibit a drop in accuracy relative to their score in Table~\ref{tab:results}, but the compositional model trained on gold decomposition still outperforms black-box models. Our error analysis of TMN-S on \datasete\ identified this key issue: While TMN-S learns to generalize, it generates questions outside the space of valid agent inputs (e.g., "Who are the directors in the movie \_\_?" vs.\ "Which movies has \_\_ directed?").

\section{Closing Remarks}

We motivated a new challenge of solving complex tasks by communicating with existing AI agents. This challenge, we believe, will help develop more generalizable and efficient models. We introduced a new benchmark dataset \dataset\ with three multi-hop reasoning challenges, all solvable by composing four QA agents. State-of-the-art language models struggle to solve \dataset, even when provided with agents' internal knowledge. 
In contrast, a model that is able to learn to communicate with the agents, albeit using annotated decompositions, is able to solve this task. These results point to the need for and the potential of such approaches, but without reliance on auxiliary annotations, to solve complex tasks.

\dataset\ is only one instantiation of our overall framework. One can extend it in many ways, such as using LMs to enrich lexical diversity,
emulating the behavior of imperfect real-world agents that even attempt to answer out-of-scope questions, diversifying to other reasoning types such as Boolean questions where using distant supervision is even harder~\cite{dasigi-etal-2019-iterative}, and extending the generalization dataset to include new examples of valid inputs as well as new agents.

\subsection*{Acknowledgments}
The authors the Aristo team at AI2, in particular Peter Clark, as well as the reviewers for valuable feedback. Experiments were conducted on beaker.org.

\bibliography{commaqa}
\bibliographystyle{acl_natbib}

\clearpage  
\input{appendix}

\end{document}

%% file: macros.tex
\newcommand{\datasetnf}{CommaQA}
\newcommand{\dataset}{\textsc{\datasetnf}}
\newcommand{\datasete}{\textsc{\datasetnf-E}}
\newcommand{\dataseti}{\textsc{\datasetnf-I}}
\newcommand{\datasetn}{\textsc{\datasetnf-N}}

\newcommand{\complext}{\ensuremath{\mathcal{T}}}
\newcommand{\complexd}{\ensuremath{\mathcal{D}}}
\newcommand{\cx}{\ensuremath{x}}
\newcommand{\cy}{\ensuremath{y}}

\newcommand{\subm}{\ensuremath{f}}
\newcommand{\submd}{{\ensuremath{\mathcal{D}_f}}}
\newcommand{\subml}{{\ensuremath{\mathcal{L}}}}
\newcommand{\submx}{\ensuremath{u}}
\newcommand{\submy}{\ensuremath{v}}

\newcommand{\kb}{\ensuremath{\mathcal{K}}}
\newcommand{\fact}{\ensuremath{\mathcal{F}}}
\newcommand{\decomp}{\ensuremath{\mathcal{D}}}

\newcommand{\opmodq}[3]{\textcolor{red}{(#1)} \textcolor{blue}{[#2]} "#3"}

\newcommand\T{\rule{0pt}{2.4ex}}

\newcommand{\selrow}{\rowcolor{violet!15}}

\newcommand{\textbox}[1]{
\noindent\fbox{%
    \parbox{0.97\linewidth}{%
      {#1}
      }%
    }%
}

%% file: appendix.tex
\appendix

\section{Multiple Answers in a Question}
\label{app:multi_ans}
If a question refers to multiple answers, e.g. "Is \#3 a part of \#2?", the operator execution is unclear. To handle such cases, the operator must specify the answer to operate over as a parameter.  E.g. \opmodq{filter(\#3)}{mathqa}{Is \#3 a part of \#2?} would filter the answers in \#3 whereas \opmodq{filter(\#2)}{mathqa}{Is \#3 a part of \#2?} would filter the answers in \#2. 

\section{Search Approach Details}
\label{app:search_model}
We describe in more detail the approach used to build the training data $\hat{\decomp}$ using the simple search technique. To generate the space of possible decompositions, for each question, we first select $f$ operations from the list of valid operations in Table~\ref{tab:operation_list}. We only consider these operations as these are the only operators needed for \dataset. Note that even with this restricted set of operators, models struggle on \dataseti\ and \datasetn. Additionally, we only consider the select operation for the first step. For all subsequent steps, we only consider replacements of \_\_ with a previous answer index. 

To select the questions, we first simplify the space of inputs by converting the questions into Fill-In-The-Blank (FITB) questions by removing the named entities. E.g "Who was born in 1991?" is changed to "Who was born in \_\_?". This is also a necessary step as the operators need questions with placeholders to handle structured answers. At every step, we expand this pool of questions by replacing the blanks with entities in the complex question and any answer index from the previous steps (e.g. \#1, \#2 in the third step of a decomposition). To avoid wasteful sampling, we use lexical overlap between questions in this expanded question pool and the input question to identify the top $g$ most relevant questions. The agent associated with each question is tracked throughout this process.

In the end, we consider the cross product between the $f$ operations and $g$ questions to produce $l=f \times g$ total questions at each steps. These $l$ questions are then executed using the appropriate agent and only the successful questions (i.e. answered by the agent) are considered for the next step. This is the key reason why the search space is much smaller than $l^o$ for $o$ reasoning steps.

\begin{table}[htbp]
    \begin{tabular}{l}
    select\\
    filter\\
    filterValues\_keys\\
    filter(\_\_) \\
    filterValues(\_\_)\_keys \\
    project \\
    projectValues \\
    projectValues\_flat \\
    projectValues\_flat\_unique \\
    project\_values\_flat \\
    project\_values\_flat\_unique \\
    \end{tabular}
    \caption{Set of operations considered in the search approach. \_\_ can be replaced by any of the answer indices from the previous steps to create a new operation.}
    \label{tab:operation_list}
\end{table}

Table~\ref{tab:search_stats} presents the overall statistics of the search approach. 
\begin{table}[htbp]
    \centering
    \includegraphics[scale=0.28]{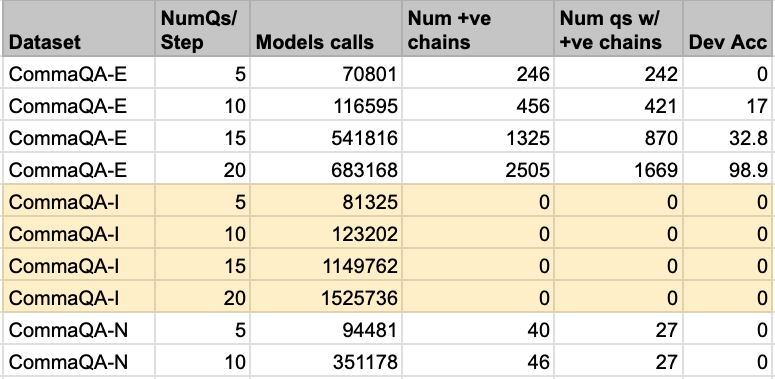}
    \caption{Statistic of the search-based approach for different values of $l$ (NumQs/Step). While we get few +ve chains for \datasetn, it is not sufficient to train an effective model.}
    \label{tab:search_stats}
\end{table}


\begin{table}[htbp]
    \centering
    \includegraphics[scale=0.4]{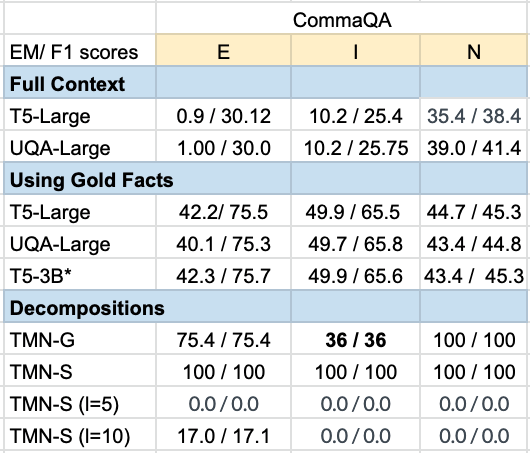}
    \caption{EM / F1 scores on the test set using the baseline approaches.}
    \label{tab:full_stats}
\end{table}

\begin{figure}[htbp]
    \centering
    \includegraphics[width=0.9\linewidth]{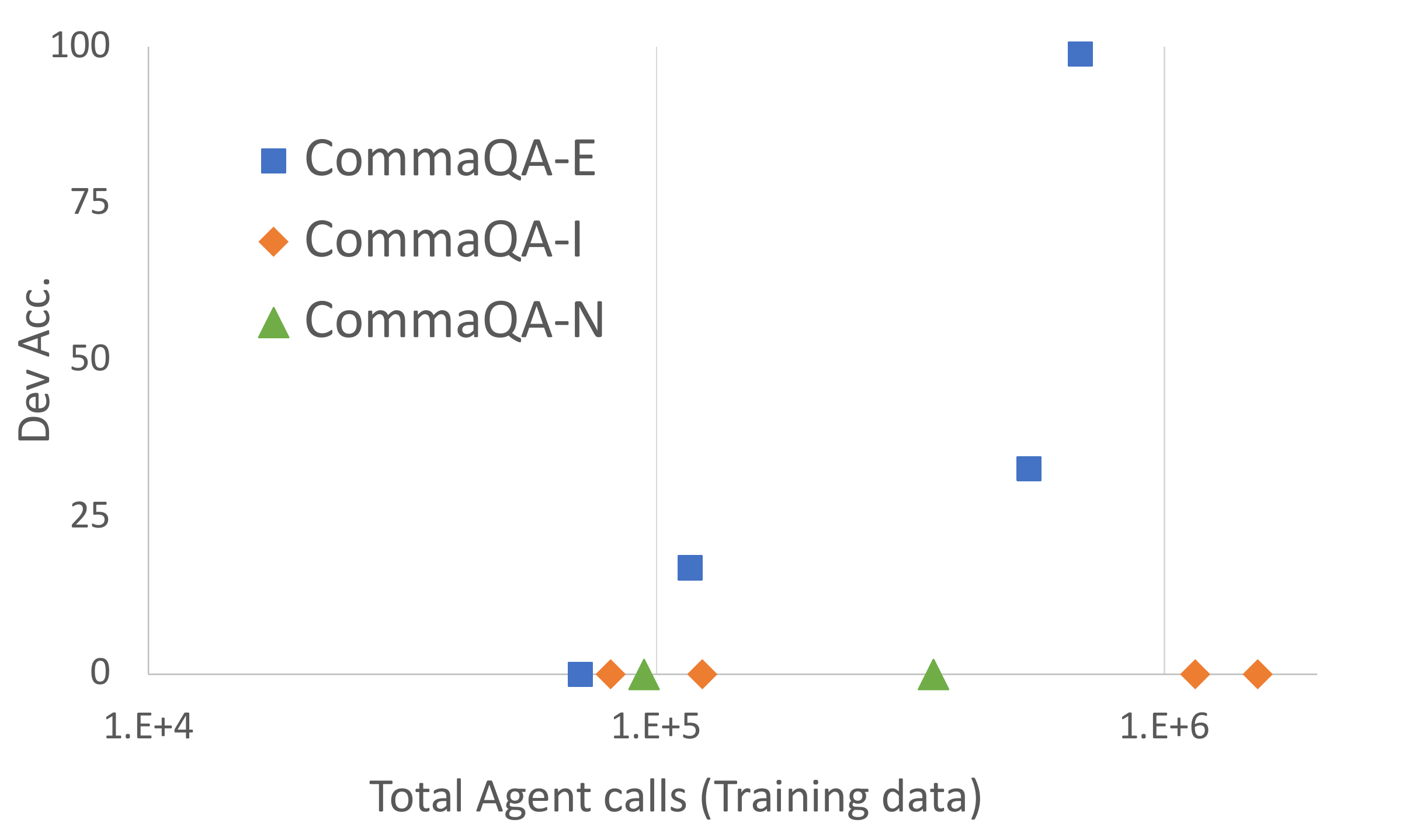}
    \caption{With an order of magnitude increase in search space, we can achieve close to 100\% accuracy on \datasete. However \dataseti\ and \datasetn\ need smarter search strategies to generate useful training supervision.
    }
    \label{fig:costvsacc}
\end{figure}

\section{Search Cost vs Accuracy}
\label{app:costvacc}
One could always exhaustively search for \emph{all} possible decompositions to reproduce the gold decompositions for all the questions. But this would be computationally highly expensive as each call to the agent would often invoke a large-scale LM or a complex AI assistant. To characterize the computational cost of these approaches, we extend the search parameter to include $l$=15 and $l$=20 (capped at 5M agent calls) and compute the accuracy of the TMN-S model trained on the resulting dataset (shown in  Fig.~\ref{fig:costvsacc}). We can achieve  close to 100\% accuracy on \datasete\ where the search is sufficiently exhaustive(about 700K model calls) mainly due to the shorter rules and the lexical signal. \dataseti\ and \datasetn, on the other hand, even with an order of magnitude increase in the number of agent calls, we don't observe any increase in the model accuracy. 
\begin{figure*}
    \centering
    \includegraphics[width=\linewidth]{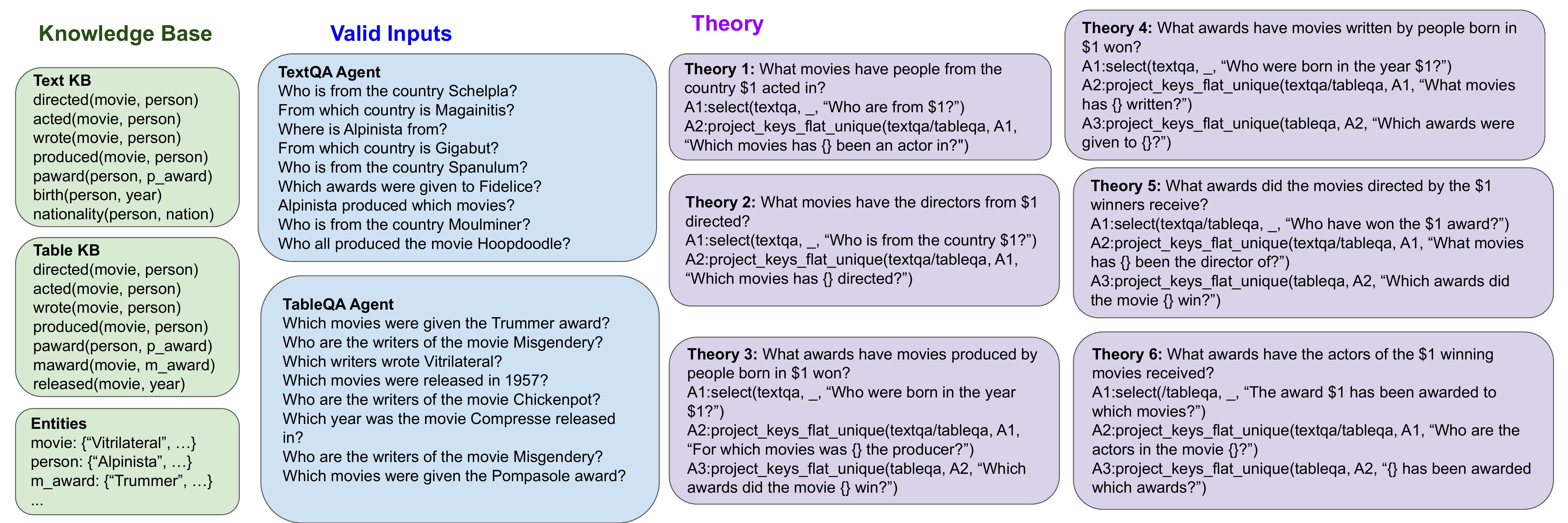}
    \caption{Example KB, space of valid inputs, and the theory used to construct \datasete.}
    \label{fig:datasete}
\end{figure*}

\section{Black-Box Models}
We train the T5 models on each of the three datasets to generate the answer given the question and facts. We format the input sequence as \texttt{<concatenated facts> Q: <question> A:}. Since many of the answers can be multiple spans, we sort\footnote{To ensure a deterministic order, we sort the answers in alphabetical order.} and concatenate them into a single string with `+' as the separator. As noted in Table~\ref{tab:dataset_stats}, the verbalized facts can result in a context over 2K tokens long. We trained T5-Large models on A100 80G GPUs and RTX8000s to train on such a long context. Transformers designed for longer documents~\cite{Beltagy2020Longformer,bigbird} would be able to handle such contexts more efficiently but generally under-perform due to sparse attention. Hence we don't evaluate them here.   

For all T5-based models, model tuning was standardly performed using a random hyper-parameter search in the style of \citet{devlin-etal-2019-bert} using the public huggingface implementation \cite{wolf2020transformers}; model selection was done based on the highest EM accuracy on the development sets. We specifically experimented with learning rates in the range of ($1e$-$3f$ to $5e$-$5f$) using both Adam and Adagrad optimizers and generally found the settings comparable to the original T5 pre-training parameters \cite{raffel2020exploring} to be optimal (Adafactor, lr=$0.001$, 10 epochs, 0-1000 warmup steps, gradient accumulation was used extensively in place of batching to fit long sequences into GPU memory). The optimal T5-3B models and T5-L for full context on \datasete\ were trained with lr=5e-5. All other models were trained with a lr of 1e-3. We will release the complete list of optimal hyper-parameters along with the code. 

\subsection{Models with Fact Supervision}
\label{app:model_fact}
To select the relevant facts, we train a RoBERTa-Large~\cite{liu2019roberta} model on the gold facts and select the top-scoring facts to produce a shorter context that fits in 512 tokens. The RoBERTa model was training using the AllenNLP library~\cite{Gardner2017AllenNLP} with the standard parameters used for RoBERTa -- learning rate of 2e-5, triangular LR scheduler with 10\% warmup steps, gradient clipping at 1.0, batch size of 16, 5 epochs of training with patience of 3 epochs. We didn't observe a noticeable difference in score with a random parameter search, so kept these parameters constant. The model was trained to score each fact independently on the train set and the best model was selected based on the accuracy on the dev set. The model was then evaluated on the facts from the train, dev and test set to produce the shorter context for all three sets. The facts were sorted based on the model's scores and the top-scoring facts were added to the context till the number of tokens did not exceed 512 tokens (white-space splitting). 

\section{Text Modular Networks: Training}
To train the NextGen model for TMNs, we use the same parameters as the prior work~\cite{khot-etal-2021-text}. We train a T5-Large model as the NextGen model using  a batch size of 64, lr of 5e-6, 5 epochs and warmup of 1000 steps in all our experiments. We used the public huggingface implementation \cite{wolf2020transformers} to train this model. During inference, we use a beam size of 10 and select 5 questions at each step. We use nucleus sampling with p=0.95 and k=10. For greedy search, we use the same parameters but select one question at each step. We use the sum log likelihood of each generated question as the score of the reasoning chain. (see released code for the exact settings) 

\begin{figure*}
    \centering
    \includegraphics[width=\textwidth]{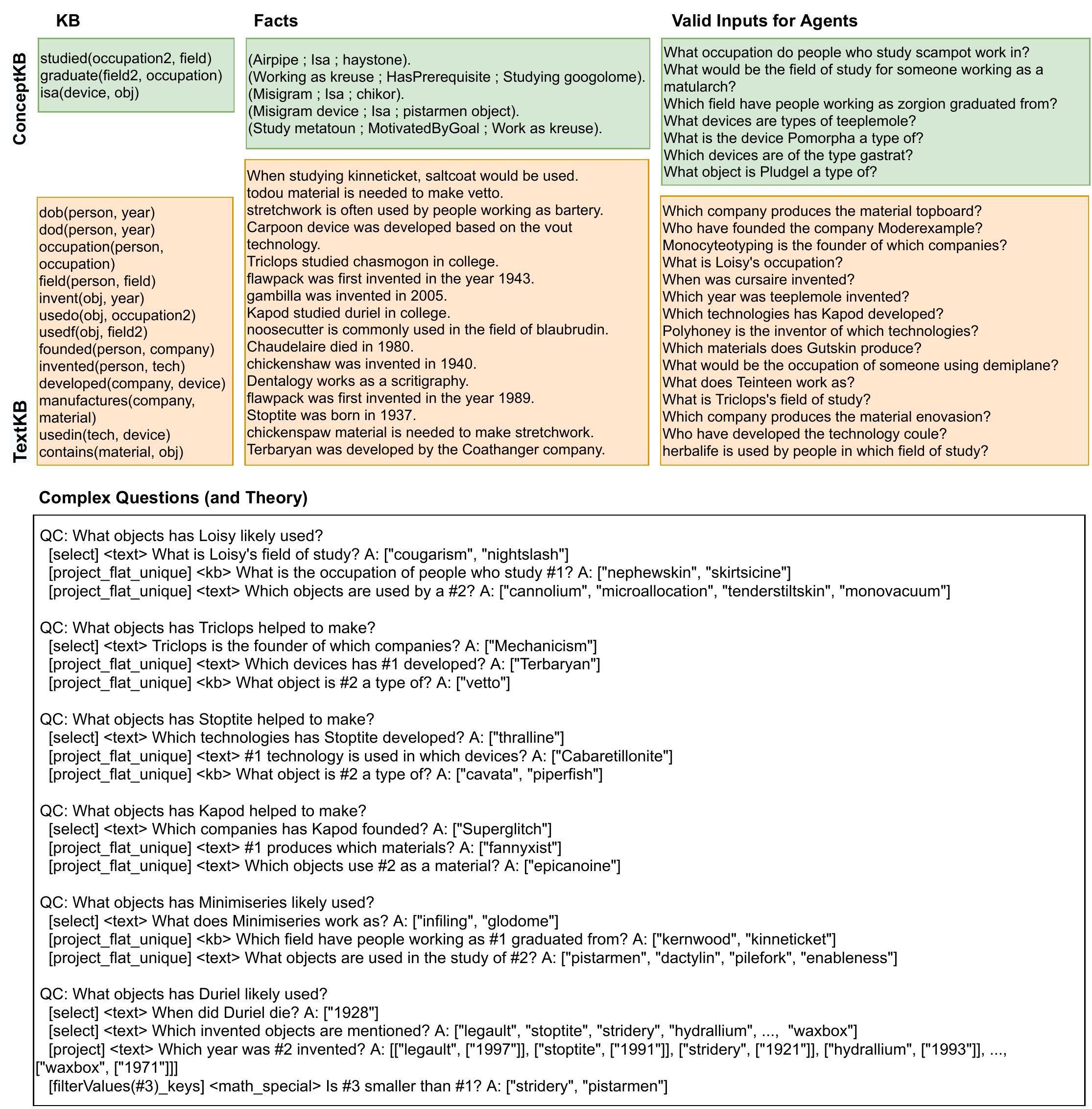}
    \caption{Example KB, space of valid inputs, and the theory used to construct \dataseti.}
    \label{fig:dataseti}
\end{figure*}

\begin{figure*}
    \centering
    \includegraphics[width=\textwidth]{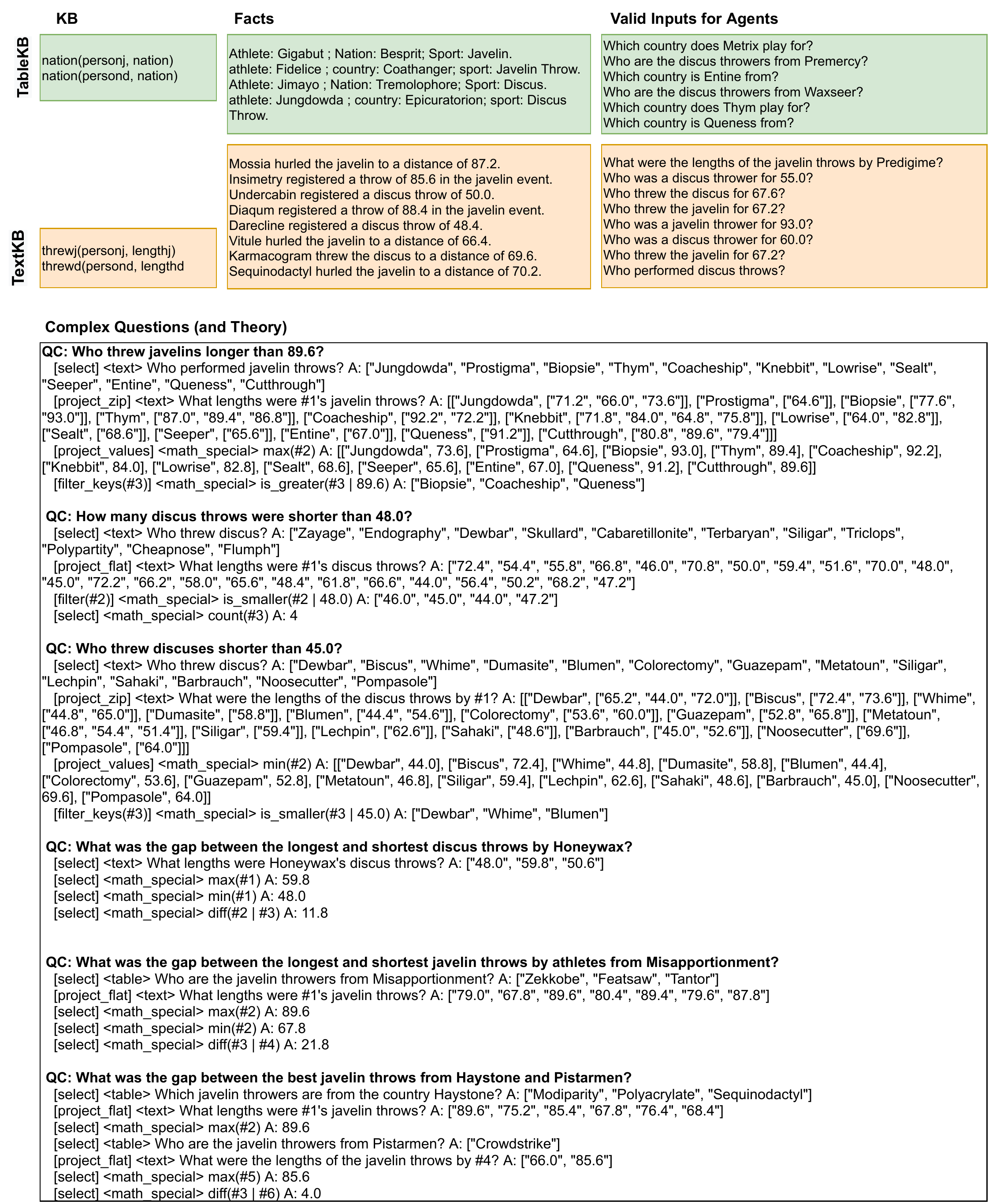}
    \caption{Example KB, space of valid inputs, and the theory used to construct \datasetn.}
    \label{fig:datasetn}
\end{figure*}

%% file: commaqa.bbl
\begin{thebibliography}{50}
\expandafter\ifx\csname natexlab\endcsname\relax\def\natexlab#1{#1}\fi

\bibitem[{Amini et~al.(2019)Amini, Gabriel, Lin, Koncel-Kedziorski, Choi, and
  Hajishirzi}]{Amini2019MathQATI}
Aida Amini, Saadia Gabriel, Shanchuan Lin, Rik Koncel-Kedziorski, Yejin Choi,
  and Hannaneh Hajishirzi. 2019.
\newblock {MathQA}: Towards interpretable math word problem solving with
  operation-based formalisms.
\newblock In \emph{NAACL}.

\bibitem[{Ammanabrolu et~al.(2021)Ammanabrolu, Urbanek, Li, Szlam,
  Rockt{\"a}schel, and Weston}]{ammanabrolu2021motivate}
Prithviraj Ammanabrolu, Jack Urbanek, Margaret Li, Arthur Szlam, Tim
  Rockt{\"a}schel, and Jason Weston. 2021.
\newblock How to motivate your dragon: Teaching goal-driven agents to speak and
  act in fantasy worlds.
\newblock In \emph{NAACL}.

\bibitem[{Beltagy et~al.(2020)Beltagy, Peters, and
  Cohan}]{Beltagy2020Longformer}
Iz~Beltagy, Matthew~E. Peters, and Arman Cohan. 2020.
\newblock Longformer: The long-document transformer.
\newblock \emph{arXiv:2004.05150}.

\bibitem[{Berant et~al.(2013)Berant, Chou, Frostig, and
  Liang}]{berant2013semantic}
Jonathan Berant, Andrew Chou, Roy Frostig, and Percy Liang. 2013.
\newblock Semantic {P}arsing on {F}reebase from {Q}uestion-{A}nswer {P}airs.
\newblock In \emph{EMNLP}.

\bibitem[{Betz and Richardson(2021)}]{betz2021deepa2}
Gregor Betz and Kyle Richardson. 2021.
\newblock {DeepA2}: A modular framework for deep argument analysis with
  pretrained neural text2text language models.
\newblock \emph{arXiv:2110.01509}.

\bibitem[{Brown et~al.(2020)Brown, Mann, Ryder, Subbiah, Kaplan, Dhariwal,
  Neelakantan, Shyam, Sastry, Askell, Agarwal, Herbert-Voss, Krueger, Henighan,
  Child, Ramesh, Ziegler, Wu, Winter, Hesse, Chen, Sigler, Litwin, Gray, Chess,
  Clark, Berner, McCandlish, Radford, Sutskever, and Amodei}]{gpt3}
Tom~B. Brown, Benjamin Mann, Nick Ryder, Melanie Subbiah, Jared Kaplan,
  Prafulla Dhariwal, Arvind Neelakantan, Pranav Shyam, Girish Sastry, Amanda
  Askell, Sandhini Agarwal, Ariel Herbert-Voss, Gretchen Krueger, T.~J.
  Henighan, Rewon Child, Aditya Ramesh, Daniel~M. Ziegler, Jeff Wu, Clemens
  Winter, Christopher Hesse, Mark Chen, Eric Sigler, Mateusz Litwin, Scott
  Gray, Benjamin Chess, Jack Clark, Christopher Berner, Sam McCandlish, Alec
  Radford, Ilya Sutskever, and Dario Amodei. 2020.
\newblock Language models are few-shot learners.
\newblock \emph{ArXiv}, abs/2005.14165.

\bibitem[{Chen et~al.(2020)Chen, Liang, Yu, Zhou, Song, and
  Le}]{Chen2020Neural}
Xinyun Chen, Chen Liang, Adams~Wei Yu, Denny Zhou, Dawn Song, and Quoc~V. Le.
  2020.
\newblock Neural symbolic reader: Scalable integration of distributed and
  symbolic representations for reading comprehension.
\newblock In \emph{ICLR}.

\bibitem[{Clark et~al.(2020)Clark, Tafjord, and
  Richardson}]{clark2020transformers}
Peter Clark, Oyvind Tafjord, and Kyle Richardson. 2020.
\newblock Transformers as soft reasoners over language.
\newblock \emph{IJCAI}.

\bibitem[{Clarke et~al.(2010)Clarke, Goldwasser, Chang, and
  Roth}]{clarke2010driving}
James Clarke, Dan Goldwasser, Ming-Wei Chang, and Dan Roth. 2010.
\newblock Driving semantic parsing from the world’s response.
\newblock In \emph{CoNLL}.

\bibitem[{Dasigi et~al.(2019)Dasigi, Gardner, Murty, Zettlemoyer, and
  Hovy}]{dasigi-etal-2019-iterative}
Pradeep Dasigi, Matt Gardner, Shikhar Murty, Luke Zettlemoyer, and Eduard Hovy.
  2019.
\newblock Iterative search for weakly supervised semantic parsing.
\newblock In \emph{NAACL-HLT}.

\bibitem[{Desai et~al.(2016)Desai, Gulwani, Hingorani, Jain, Karkare, Marron,
  and Roy}]{desai2016program}
Aditya Desai, Sumit Gulwani, Vineet Hingorani, Nidhi Jain, Amey Karkare, Mark
  Marron, and Subhajit Roy. 2016.
\newblock Program synthesis using natural language.
\newblock In \emph{ICSE}.

\bibitem[{Devlin et~al.(2019)Devlin, Chang, Lee, and
  Toutanova}]{devlin-etal-2019-bert}
Jacob Devlin, Ming-Wei Chang, Kenton Lee, and Kristina Toutanova. 2019.
\newblock {BERT}: Pre-training of deep bidirectional transformers for language
  understanding.
\newblock In \emph{NAACL}.

\bibitem[{Dua et~al.(2019)Dua, Wang, Dasigi, Stanovsky, Singh, and
  Gardner}]{Dua2019DROP}
Dheeru Dua, Yizhong Wang, Pradeep Dasigi, Gabriel Stanovsky, Sameer Singh, and
  Matt Gardner. 2019.
\newblock {DROP}: A reading comprehension benchmark requiring discrete
  reasoning over paragraphs.
\newblock In \emph{NAACL}.

\bibitem[{Fried et~al.(2015)Fried, Jansen, Hahn-Powell, Surdeanu, and
  Clark}]{Fried2015HigherorderLS}
Daniel Fried, Peter~A. Jansen, Gus Hahn-Powell, Mihai Surdeanu, and Peter~E.
  Clark. 2015.
\newblock Higher-order lexical semantic models for non-factoid answer
  reranking.
\newblock \emph{TACL}, 3:197--210.

\bibitem[{Gardner et~al.(2019)Gardner, Berant, Hajishirzi, Talmor, and
  Min}]{Gardner2019QuestionAI}
Matt Gardner, Jonathan Berant, Hannaneh Hajishirzi, Alon Talmor, and Sewon Min.
  2019.
\newblock Question answering is a format; when is it useful?
\newblock \emph{ArXiv}, abs/1909.11291.

\bibitem[{Gardner et~al.(2017)Gardner, Grus, Neumann, Tafjord, Dasigi, Liu,
  Peters, Schmitz, and Zettlemoyer}]{Gardner2017AllenNLP}
Matt Gardner, Joel Grus, Mark Neumann, Oyvind Tafjord, Pradeep Dasigi,
  Nelson~F. Liu, Matthew Peters, Michael Schmitz, and Luke~S. Zettlemoyer.
  2017.
\newblock {AllenNLP}: A deep semantic natural language processing platform.
\newblock \emph{arXiv preprint arXiv:1803.07640}.

\bibitem[{Geva et~al.(2021)Geva, Khashabi, Segal, Khot, Roth, and
  Berant}]{geva2021strategyqa}
Mor Geva, Daniel Khashabi, Elad Segal, Tushar Khot, Dan Roth, and Jonathan
  Berant. 2021.
\newblock {Did Aristotle Use a Laptop? A Question Answering Benchmark with
  Implicit Reasoning Strategies}.
\newblock \emph{TACL}.

\bibitem[{Gulwani(2011)}]{gulwani2011automating}
Sumit Gulwani. 2011.
\newblock Automating string processing in spreadsheets using input-output
  examples.
\newblock \emph{ACM Sigplan Notices}, 46(1):317--330.

\bibitem[{Gupta et~al.(2021)Gupta, Kamath, Kembhavi, and
  Hoiem}]{Gupta2021TowardsGP}
Tanmay Gupta, Amita Kamath, Aniruddha Kembhavi, and Derek Hoiem. 2021.
\newblock Towards general purpose vision systems.
\newblock \emph{ArXiv}, abs/2104.00743.

\bibitem[{Hausknecht et~al.(2020)Hausknecht, Ammanabrolu, C{\^o}t{\'e}, and
  Yuan}]{hausknecht2020interactive}
Matthew Hausknecht, Prithviraj Ammanabrolu, Marc-Alexandre C{\^o}t{\'e}, and
  Xingdi Yuan. 2020.
\newblock Interactive fiction games: A colossal adventure.
\newblock In \emph{AAAI}.

\bibitem[{Ho et~al.(2020)Ho, Nguyen, Sugawara, and
  Aizawa}]{xanh2020_2wikimultihop}
Xanh Ho, A.~Nguyen, Saku Sugawara, and Akiko Aizawa. 2020.
\newblock Constructing a multi-hop qa dataset for comprehensive evaluation of
  reasoning steps.
\newblock In \emph{COLING}.

\bibitem[{Jansen(2021)}]{jansen2021systematic}
Peter~A Jansen. 2021.
\newblock A systematic survey of text worlds as embodied natural language
  environments.
\newblock \emph{arXiv preprint arXiv:2107.04132}.

\bibitem[{Khashabi et~al.(2019)Khashabi, Azer, Khot, Sabharwal, and
  Roth}]{Khashabi2019OnTP}
Daniel Khashabi, Erfan~Sadeqi Azer, Tushar Khot, Ashish Sabharwal, and Dan
  Roth. 2019.
\newblock On the possibilities and limitations of multi-hop reasoning under
  linguistic imperfections.
\newblock \emph{arXiv}, abs/1901.02522.

\bibitem[{Khashabi et~al.(2018)Khashabi, Chaturvedi, Roth, Upadhyay, and
  Roth}]{MultiRC2018}
Daniel Khashabi, Snigdha Chaturvedi, Michael Roth, Shyam Upadhyay, and Dan
  Roth. 2018.
\newblock Looking beyond the surface:a challenge set for reading comprehension
  over multiple sentences.
\newblock In \emph{NAACL}.

\bibitem[{Khashabi et~al.(2020)Khashabi, Min, Khot, Sabhwaral, Tafjord, Clark,
  and Hajishirzi}]{2020unifiedqa}
Daniel Khashabi, Sewon Min, Tushar Khot, Ashish Sabhwaral, Oyvind Tafjord,
  Peter Clark, and Hannaneh Hajishirzi. 2020.
\newblock {UnifiedQA}: Crossing format boundaries with a single {QA} system.
\newblock In \emph{Findings of EMNLP}.

\bibitem[{Khot et~al.(2020)Khot, Clark, Guerquin, Jansen, and Sabharwal}]{qasc}
Tushar Khot, Peter Clark, Michal Guerquin, Paul~Edward Jansen, and Ashish
  Sabharwal. 2020.
\newblock {QASC}: A dataset for question answering via sentence composition.
\newblock In \emph{AAAI}.

\bibitem[{Khot et~al.(2021)Khot, Khashabi, Richardson, Clark, and
  Sabharwal}]{khot-etal-2021-text}
Tushar Khot, Daniel Khashabi, Kyle Richardson, Peter Clark, and Ashish
  Sabharwal. 2021.
\newblock Text modular networks: Learning to decompose tasks in the language of
  existing models.
\newblock In \emph{NAACL}.

\bibitem[{Krishnamurthy et~al.(2017)Krishnamurthy, Dasigi, and
  Gardner}]{Krishnamurthy2017NeuralSP}
Jayant Krishnamurthy, Pradeep Dasigi, and Matt Gardner. 2017.
\newblock Neural semantic parsing with type constraints for semi-structured
  tables.
\newblock In \emph{EMNLP}.

\bibitem[{Lake and Baroni(2018)}]{lake2018generalization}
Brenden Lake and Marco Baroni. 2018.
\newblock Generalization without systematicity: On the compositional skills of
  sequence-to-sequence recurrent networks.
\newblock In \emph{ICML}, pages 2873--2882.

\bibitem[{Lewis et~al.(2021)Lewis, Stenetorp, and Riedel}]{Lewis2021TrainTest}
Patrick Lewis, Pontus Stenetorp, and Sebastian Riedel. 2021.
\newblock Question and answer test-train overlap in open-domain question
  answering datasets.
\newblock In \emph{EACL}.

\bibitem[{Liu et~al.(2021)Liu, Lee, Jia, and Liang}]{liu2021can}
Nelson~F Liu, Tony Lee, Robin Jia, and Percy Liang. 2021.
\newblock Can small and synthetic benchmarks drive modeling innovation? {A}
  retrospective study of question answering modeling approaches.
\newblock \emph{arXiv preprint arXiv:2102.01065}.

\bibitem[{Liu et~al.(2019)Liu, Ott, Goyal, Du, Joshi, Chen, Levy, Lewis,
  Zettlemoyer, and Stoyanov}]{liu2019roberta}
Yinhan Liu, Myle Ott, Naman Goyal, Jingfei Du, Mandar Joshi, Danqi Chen, Omer
  Levy, Mike Lewis, Luke Zettlemoyer, and Veselin Stoyanov. 2019.
\newblock {RoBERTa}: A robustly optimized bert pretraining approach.
\newblock \emph{arXiv:1907.11692}.

\bibitem[{Mihaylov et~al.(2018)Mihaylov, Clark, Khot, and Sabharwal}]{obqa}
Todor Mihaylov, Peter Clark, Tushar Khot, and Ashish Sabharwal. 2018.
\newblock Can a suit of armor conduct electricity? a new dataset for open book
  question answering.
\newblock In \emph{EMNLP}.

\bibitem[{Min et~al.(2019{\natexlab{a}})Min, Wallace, Singh, Gardner,
  Hajishirzi, and Zettlemoyer}]{min2019compositional}
Sewon Min, Eric Wallace, Sameer Singh, Matt Gardner, Hannaneh Hajishirzi, and
  Luke Zettlemoyer. 2019{\natexlab{a}}.
\newblock Compositional questions do not necessitate multi-hop reasoning.
\newblock In \emph{ACL}.

\bibitem[{Min et~al.(2019{\natexlab{b}})Min, Zhong, Zettlemoyer, and
  Hajishirzi}]{decomprc}
Sewon Min, Victor Zhong, Luke~S. Zettlemoyer, and Hannaneh Hajishirzi.
  2019{\natexlab{b}}.
\newblock Multi-hop reading comprehension through question decomposition and
  rescoring.
\newblock In \emph{ACL}.

\bibitem[{Perez et~al.(2020)Perez, Lewis, Yih, Cho, and
  Kiela}]{perez2020unsupervised}
Ethan Perez, Patrick Lewis, Wen-tau Yih, Kyunghyun Cho, and Douwe Kiela. 2020.
\newblock Unsupervised question decomposition for question answering.
\newblock In \emph{EMNLP}.

\bibitem[{Radford et~al.(2019)Radford, Wu, Child, Luan, Amodei, and
  Sutskever}]{gpt2}
Alec Radford, Jeff Wu, Rewon Child, David Luan, Dario Amodei, and Ilya
  Sutskever. 2019.
\newblock Language models are unsupervised multitask learners.

\bibitem[{Raffel et~al.(2020)Raffel, Shazeer, Roberts, Lee, Narang, Matena,
  Zhou, Li, and Liu}]{raffel2020exploring}
Colin Raffel, Noam Shazeer, Adam Roberts, Katherine Lee, Sharan Narang, Michael
  Matena, Yanqi Zhou, Wei Li, and Peter~J Liu. 2020.
\newblock Exploring the limits of transfer learning with a unified text-to-text
  transformer.
\newblock \emph{Journal of Machine Learning Research}, 21:1--67.

\bibitem[{Sinha et~al.(2019)Sinha, Sodhani, Dong, Pineau, and
  Hamilton}]{sinha2019clutrr}
Koustuv Sinha, Shagun Sodhani, Jin Dong, Joelle Pineau, and William~L Hamilton.
  2019.
\newblock {CLUTRR}: A diagnostic benchmark for inductive reasoning from text.
\newblock In \emph{EMNLP}.

\bibitem[{Talmor and Berant(2018)}]{talmor2018web}
Alon Talmor and Jonathan Berant. 2018.
\newblock The web as a knowledge-base for answering complex questions.
\newblock In \emph{NAACL}.

\bibitem[{Trivedi et~al.(2020)Trivedi, Balasubramanian, Khot, and
  Sabharwal}]{Trivedi2020DiRe}
Harsh Trivedi, Niranjan Balasubramanian, Tushar Khot, and Ashish Sabharwal.
  2020.
\newblock Is multihop {QA} in {DiRe} condition? {M}easuring and reducing
  disconnected reasoning.
\newblock In \emph{EMNLP}.

\bibitem[{Trivedi et~al.(2021)Trivedi, Balasubramanian, Khot, and
  Sabharwal}]{musique}
Harsh Trivedi, Niranjan Balasubramanian, Tushar Khot, and Ashish Sabharwal.
  2021.
\newblock {MuSiQue}: Multi-hop questions via single-hop question composition.
\newblock \emph{ArXiv}, abs/2108.00573.

\bibitem[{Wang et~al.(2019)Wang, Pruksachatkun, Nangia, Singh, Michael, Hill,
  Levy, and Bowman}]{superglue}
Alex Wang, Yada Pruksachatkun, Nikita Nangia, Amanpreet Singh, Julian Michael,
  Felix Hill, Omer Levy, and Samuel Bowman. 2019.
\newblock Superglue: A stickier benchmark for general-purpose language
  understanding systems.
\newblock In \emph{NeurIPS}, volume~32.

\bibitem[{Wang et~al.(2018)Wang, Singh, Michael, Hill, Levy, and Bowman}]{glue}
Alex Wang, Amanpreet Singh, Julian Michael, Felix Hill, Omer Levy, and Samuel
  Bowman. 2018.
\newblock {GLUE}: A multi-task benchmark and analysis platform for natural
  language understanding.
\newblock In \emph{Blackbox NLP Workshop}.

\bibitem[{Wolf et~al.(2020)Wolf, Chaumond, Debut, Sanh, Delangue, Moi, Cistac,
  Funtowicz, Davison, Shleifer et~al.}]{wolf2020transformers}
Thomas Wolf, Julien Chaumond, Lysandre Debut, Victor Sanh, Clement Delangue,
  Anthony Moi, Pierric Cistac, Morgan Funtowicz, Joe Davison, Sam Shleifer,
  et~al. 2020.
\newblock Transformers: State-of-the-art natural language processing.
\newblock In \emph{EMNLP: System Demonstrations}.

\bibitem[{Wolfson et~al.(2020)Wolfson, Geva, Gupta, Gardner, Goldberg, Deutch,
  and Berant}]{Wolfson2020Break}
Tomer Wolfson, Mor Geva, Ankit Gupta, Matt Gardner, Yoav Goldberg, Daniel
  Deutch, and Jonathan Berant. 2020.
\newblock Break it down: {A} question understanding benchmark.
\newblock \emph{TACL}.

\bibitem[{Yang et~al.(2018)Yang, Qi, Zhang, Bengio, Cohen, Salakhutdinov, and
  Manning}]{hotpotqa}
Zhilin Yang, Peng Qi, Saizheng Zhang, Yoshua Bengio, William~W. Cohen, Ruslan
  Salakhutdinov, and Christopher~D. Manning. 2018.
\newblock {HotpotQA}: A dataset for diverse, explainable multi-hop question
  answering.
\newblock In \emph{EMNLP}.

\bibitem[{Yuan et~al.(2019)Yuan, C{\^o}t{\'e}, Fu, Lin, Pal, Bengio, and
  Trischler}]{yuan2019interactive}
Xingdi Yuan, Marc-Alexandre C{\^o}t{\'e}, Jie Fu, Zhouhan Lin, Christopher Pal,
  Yoshua Bengio, and Adam Trischler. 2019.
\newblock Interactive language learning by question answering.
\newblock In \emph{EMNLP-IJCNLP}.

\bibitem[{Yuan et~al.(2020)Yuan, Fu, C{\^o}t{\'e}, Tay, Pal, and
  Trischler}]{yuan2020interactive}
Xingdi Yuan, Jie Fu, Marc-Alexandre C{\^o}t{\'e}, Yi~Tay, Christopher Pal, and
  Adam Trischler. 2020.
\newblock Interactive machine comprehension with information seeking agents.
\newblock In \emph{ACL}.

\bibitem[{Zaheer et~al.(2020)Zaheer, Guruganesh, Dubey, Ainslie, Alberti,
  Ontanon, Pham, Ravula, Wang, Yang et~al.}]{bigbird}
Manzil Zaheer, Guru Guruganesh, Kumar~Avinava Dubey, Joshua Ainslie, Chris
  Alberti, Santiago Ontanon, Philip Pham, Anirudh Ravula, Qifan Wang, Li~Yang,
  et~al. 2020.
\newblock {Big Bird}: Transformers for longer sequences.
\newblock In \emph{NeurIPS}.

\end{thebibliography}
